\documentclass[times,10pt]{article}
\usepackage[letterpaper, total={6.5in, 9in}]{geometry}
\usepackage{graphicx}
\usepackage{amsmath,amssymb}
\usepackage{algorithm, algpseudocode}
\newtheorem{theorem}{Theorem}
\newtheorem{lemma}{Lemma}
\def\textsum{\mathop{\textstyle\sum}\nolimits}
\def\textprod{\mathop{\textstyle\prod}\nolimits}

\title{Dynamical System Optimization}
\author{Emo Todorov\footnote{The author is with Roboti LLC and the University of Washington. Email: etodorov@gmail.com}}

\begin{document}

\maketitle

\begin{abstract}
We develop an optimization framework centered around a core idea: once a (parametric) policy is specified, control authority is transferred to the policy, resulting in an autonomous dynamical system. Thus we should be able to optimize policy parameters without further reference to controls or actions, and without directly using the machinery of approximate Dynamic Programming and Reinforcement Learning. Here we derive simpler algorithms at the autonomous system level, and show that they compute the same quantities as policy gradients and Hessians, natural gradients, proximal methods. Analogs to approximate policy iteration and off-policy learning are also available. Since policy parameters and other system parameters are treated uniformly, the same algorithms apply to behavioral cloning, mechanism design, system identification, learning of state estimators. Tuning of generative AI models is not only possible, but is conceptually closer to the present framework than to Reinforcement Learning.

\end{abstract}

\section{Introduction}

We introduce a framework for dynamical system optimization\footnote{This is a very broad concept. Loosely-related prior work likely exists in multiple fields of science and engineering. DSO as defined here is a concrete formalism yielding novel technical results. We leave a general literature review for the future.} which we call DSO. It is simple, yet multiple problems can be expressed in it and tackled with the same machinery. They include policy optimization, behavioral cloning, mechanism design, system identification, learning of state estimators and ``brain-body dynamics'', tuning of generative AI models. Frameworks aiming at this level of generality are often limited to derivative-free (black-box) optimization. What makes DSO unique is that it combines generality with efficient optimization, retaining the essence of policy gradient and related methods \cite{sutton1999,kakade2001,silver2014,schulman2015,schulman2017,abdolmaleki2018,rajeswaran2018}.

Each DSO problem is defined by two key ingredients: a Markov chain $P(x'|x,\theta)$ and a step cost\footnote{Cost and Reward can be used interchangeably. We use Cost throughout.} $L(x, \theta)$. The reason for sharing the parameters $\theta$ will soon become clear. We then construct an objective $J(\theta)$ which has the semantics of cumulative cost, compute its gradient $\nabla_\theta J(\theta)$ and apply gradient descent; with various refinements. In the course of optimization, the Markov chain is modified so as to visit states that have low cost; while the cost function is modified so as to make visited states less costly.

While DSO aims to solve control-related problems, it has no notion of actions or decision-making over time. Indeed it is not a Markov Decision Process (MDP), but only a Markov Process or Markov chain, with a general optimization problem wrapped around it. How can it solve control problems, and specifically policy optimization? This is possible because once an MDP policy is defined, actions are no longer free decision variables, but rather intermediate quantities which are generated by the policy and which in turn affect the dynamics and cost. So it is the policy that ultimately affects the dynamics and cost. And since the policy depends on the parameters, the latter are shared between the Markov chain and the cost function.

DSO's treatment of parameters as the primary decision variables is in contrast with approximate Dynamic Programming \cite{bertsekas1996} and Reinforcement Learning \cite{sutton2018} -- where actions are the primary decision variables chosen independently at each state, while parameters are brought in later at an algorithmic stage in order to fit underlying non-parametric quantities via function approximation. We can apply function approximation in DSO as well, but only over the state space. State-action value functions ($Q$ functions) cannot be defined, which means that off-policy $Q$-learning cannot be applied. However, in our earlier work on Linearly-solvable MDPs (LMDPs) we developed a method called $Z$-learning \cite{todorov2006}, which yields off-chain learning in DSO.

The rest of the paper is organized as follows. In Section 2 we define the DSO objective, and show that policy optimization for different families of MDPs can be expressed as an equivalent DSO problem. We also show how pairs of MDPs that map to the same DSO problem can be converted into each other. In Section 3 we derive the DSO gradient, and show that MDP policy gradient theorems for stochastic \cite{sutton1999} and deterministic \cite{silver2014} policies as well as LDMP policies \cite{todorov2010} can be recovered from this unified quantity. In section 4 we develop the DSO analogs to proximal methods, natural gradients, Hessian estimation, and variance reduction methods. We also identify a surrogate objective which has the same gradient, and which facilitates sampling approximations and yields an analog to policy iteration. The LMDP formulation is transferred to the DSO framework, enabling off-chain learning. In Section 5 we explain how problems beyond policy optimization can be expressed in DSO.

An ``actionable summary'' of this paper would include DSO construction in Section 2.1 (page 3), DSO gradient estimation in Algorithm 1 (page 11), and application notes in Section 5 (page 17).

\section{Dynamical system optimization (DSO)}

Let $x \in \mathcal{R}^{n_x}$ be the state of a Markov chain with transition probability $P(x'|x, \theta)$ and step cost $L(x, \theta)$, both parameterized by the same vector $\theta \in \mathcal{R}^{n_\theta}$. We seek to optimize an objective $J(\theta)$ which has the semantics of cumulative cost; defined below in episodic, average, and time-varying settings. In each setting we also define the value function $V(x,\theta)$ which measures long-term performance starting from a given state. This function satisfies simplified Bellman equations. The definitions closely resemble the corresponding quantities in MDPs \cite{bertsekas1996,sutton2018} except they are simplified by removing any reference to actions.

We will be differentiating with respect to $\theta$ so it needs to be continuous, however $x$ can be made discrete and our results still hold -- by replacing integrals with sums. We will be changing the order of gradients and integration, requiring mild regularity conditions (which we assume.) When actions $a \in \mathcal{R}^{n_a}$ are introduced later we will treat them as continuous, but they can also be made discrete (except as noted.)

\subsubsection*{Episodic cost}

Here the initial state is sampled from some initial state distribution $P_0(x)$ and costs are accumulated over an episode. In an infinite-horizon formulation, the episode continues indefinitely and costs are discounted exponentially over time with discount factor $\gamma \in (0,1)$. In a first-exit formulation, the episode ends when a terminal state $x \in \mathcal{T}$ is first reached. In that case we can set $\gamma = 1$, but only when the chain is guaranteed (with probability $1$) to reach a terminal state in finite time, starting from any state.

The objective is the value function averaged over the initial state distribution:
\begin{equation}\label{EPIperf}
J(\theta) \equiv \mathbb{E}_{x \sim P_0(\cdot)} \left[ V(x, \theta) \right]
\end{equation}
where the value function is the discounted cost expected to accumulate starting at a given state:
\begin{equation}\label{EPI_V}
V(x, \theta) \equiv \mathbb{E}_{x_0=x, \thinspace x_{t+1} \sim P(\cdot|x_t,\theta)} \left[  \textsum_{t=0}^{x_t\in\mathcal{T}} \gamma^t L(x_t, \theta) \right]
\end{equation}
In the infinite-horizon formulation where the set of terminal states $\mathcal{T}$ is empty, the termination condition in the upper limit is never satisfied and so that limit is effectively $\infty$. 

A key property of the value function is that it is the (unique) solution to the Bellman equation:
\begin{equation}\label{EPIbel}
V(x,\theta) = L(x, \theta) + \gamma \mathbb{E}_{x' \sim P(\cdot|x, \theta)} \left[ V(x',\theta)\right]
\end{equation}

\noindent Another quantity that plays an important role in episodic settings is the ``discounted density'':
\begin{equation}\label{rho}
\rho(x, \theta) \equiv \int P_0(y) \left( \textsum_{t = 1}^{\infty} { \gamma^{t-1} P(y \rightarrow x, t, \theta)} \right) dy
\end{equation}
Here $P(y \rightarrow x, t, \theta)$ is the probability of transitioning from state $y$ to state $x$ after $t$ steps, taking into account any terminal stares. When $\gamma = 1$, $\rho$ is the visitation density (it is not a stationary density.) When $\gamma < 0$, $\rho$ is proportional to the visitation density of a Markov chain which at every step either proceeds according to $P(x'|x,\theta)$ with probability $\gamma$; or jumps to an abstract terminal state with probability $1-\gamma$.

In the rest of the paper we mostly focus on the infinite-horizon flavor of episodic settings and write $\infty$ instead of $x \in \mathcal{T}$, as is common in the MDP literature. But keep in mind that first-exit is also covered.

\subsubsection*{Average cost}

Here we accumulate costs indefinitely without discounting. Total cost can no longer be used as an objective because it is infinite. Instead, the objective is defined as average cost per step:
\begin{equation}\label{AVperf}
J(\theta) \equiv \mathbb{E}_{x_0 \sim P_0(\cdot), \thickspace x_{t+1} \sim P(\cdot|x_t, \theta)} \left[ \lim_{T \rightarrow \infty} \tfrac{1}{T} \textsum_{t=0}^{T} L(x_t, \theta)\right]
\end{equation}

\noindent This setting is only applicable when the Markov chain is ergodic. Let $d(x,\theta)$ denote the stationary density, which has the defining property:
\begin{equation}\label{d}
d(x,\theta) = \int P(y|x,\theta) d(y,\theta) dy
\end{equation}
Ergodic processes ``forget'' their initial state and averages only depend on the stationary density, thus:
\begin{equation}
J(\theta) = \mathbb{E}_{x \sim d(\cdot, \theta)} \left[ L(x,\theta) \right]
\end{equation}

\noindent The definition of the value function in this setting is more delicate. A straightforward definition similar to the discounted setting would yield a constant $V(x,\theta) = J(\theta)$ for all states. Instead we define the ``differential'' value function as the (unique) solution to the Bellman equation:
\begin{equation}\label{AVbel}
J(\theta) + V(x, \theta) = L(x,\theta) + \mathbb{E}_{x' \sim P(\cdot|x,\theta)} \left[ V(x', \theta) \right]   
\end{equation}

\subsubsection*{Time-varying cost}

Here we accumulate costs over a predefined time horizon $T$ and allow both the cost function $L_t(x,\theta)$ and the Markov chain $P_t(x'|x,\theta)$ to be time-varying; while the parameters $\theta$ remain time-invariant. Discounting can be incorporated into $L_t$ if desired. The value function now also becomes time-varying:
\begin{equation}\label{TIME_V}
V_t(x,\theta) \equiv \mathbb{E}_{x_t = x, \thinspace x_{k+1} \sim P_k(\cdot|k_t, \theta)} \left[ \textsum_{k=t}^{T} L_k(x_k,\theta)\right]
\end{equation}
Similar to the episodic setting, the objective is the value averaged over the initial state distribution:
\begin{equation}\label{TIMEperf}
J(\theta) \equiv \mathbb{E}_{x \sim P_0(\cdot, \theta)} \left[ V_0(x,\theta) \right]
\end{equation}
This is technically an episodic setting: we can augment the state with the time index and define a set of terminal states accordingly. But we are treating it separately because the corresponding algorithms are different; in particular, function approximation makes little sense for time-varying value functions (unless the approximation is a quadratic.) The Bellman equation is now an explicit recursion over time:
\begin{equation}\label{TIMEbel}
V_t(x,\theta) = L_t(x,\theta) + \mathbb{E}_{x' \sim P_t(\cdot|x, \theta)} \left[ V_{t+1}(x',\theta) \right]
\end{equation}
At the last time step we have $V_T(x,\theta) = L_T(x,\theta)$.

\subsection{MDP policy optimization as a DSO problem}

We now show how policy optimization in existing families of MDPs can be expressed as an equivalent DSO problem. The idea is to construct a Markov chain $P(x'|x,\theta)$ which is guaranteed to have the same transition probabilities as the controlled MDP; and cost function $L(x,\theta)$ such that the Markov chain is guaranteed to incur the same cumulative cost as the controlled MDP starting from the same state; meaning that the MDP and the DSO have the same state value function. Note that the DSO cost itself cannot match the MDP cost, because the latter depends on actions which are absent in DSO. But since MDP policy performance depends on the cost only through the value function, matching value functions (in addition to transition probabilities) is sufficient. To avoid notational confusion, we use lower-case symbols for MDP-specific quantities.

We handle multiple MDP families uniformly by first defining a general MDP, mapping it to DSO, and then showing how existing MDP families are special cases. This general MDP has state $x$, action $a$, transition probability $p(x'|x, a)$, parametric policy $\pi(a|x,\theta)$, and step cost $\ell(x,a,\theta)$ allowed to depend on the same parameters as the policy. The latter dependence is unusual, but we will need it to capture various forms of (cross) entropy regularization used in the literature. This general MDP will be called G-MDP.

\textbf{G-MDP} is mapped to DSO as:
\begin{equation}\label{GMDP}
\begin{aligned}
P_G(x'|x,\theta) & \equiv \int \pi(a|x,\theta) \thinspace p(x'|x,a) \thinspace da \\
L_G(x,\theta) & \equiv \int \pi(a|x,\theta) \thinspace \ell(x,a,\theta) \thinspace da
\end{aligned}
\end{equation}
We will prove that G-MDP and the above DSO are equivalent. The Markov chain $P_G$ equals the controlled transition probabilities in G-MDP by construction. Equality of the value functions is established by the following lemma.

\begin{lemma}[MDP policy evaluation via Markov chain] Let $v(x,\theta)$ denote the (episodic or average) value function for an MDP with transition probability $p(x'|x,a)$ and cost $\ell(x,a,\theta)$ under policy $\pi(a|x, \theta)$. Let $L_G(x,\theta)$ and $P_G(x'|x,\theta)$ be defined as in (\ref{GMDP}). Then in the episodic setting $v(x,\theta)$ satisfies:
\begin{equation}
v(x, \theta) = L_G(x, \theta) + \gamma \int P_G(x'|x, \theta) v(x', \theta) dx'
\end{equation}
In the average setting, defined only when $P_G(x'|x,\theta)$ is ergodic, $v(x,\theta)$ satisfies:
\begin{equation}\label{StocBel}
J(\theta) + v(x, \theta) = L_G(x, \theta) + \int P_G(x'|x, \theta) v(x', \theta) dx'
\end{equation}
where $J(\theta) = \mathbb{E}_{x \sim d(\cdot, \theta)} \left[L(x,\theta)\right]$ is the average cost per step and $d(x,\theta)$ is the stationary density.
\end{lemma}

\noindent \textbf{Proof.} The Bellman equation for the policy-specific episodic value function in an MDP is:
\begin{equation}
v(x, \theta) = \mathbb{E}_{a \sim \pi(\cdot| x, \theta)} \left[ \ell(x,a,\theta) + \gamma \mathbb{E}_{x' \sim p(\cdot|x, a)} \left[ v(x', \theta) \right] \right]
\end{equation}
Writing the expectations as integrals and grouping terms yields:
\begin{equation}
v(x, \theta) = \underbrace{\int \pi(a|x,\theta) \ell(x,a,\theta) da}_{ L_G(x, \theta)} + \gamma \int \underbrace{\left( \int \pi(a|x,\theta) p(x'|x,a)  da \right)}_{P_G(x'|x,\theta)} v(x', \theta) \thinspace dx'
\end{equation}
where we have recognized the definitions from (\ref{GMDP}). The proof for the average case is identical. $\square$

Thus the G-MDP value function $v(x,\theta)$ and the corresponding DSO value function $V(x,\theta)$ satisfy the same Bellman equation in both the episodic and average settings. And since the Bellman equation has a unique solution, we have established that $v(x,\theta) = V(x,\theta)$ for all $x,\theta$. We could further make the G-MDP cost stochastic, and replace $\ell(x,a,\theta)$ with its expectation  in (\ref{GMDP}), but this is not needed here.

Next we explain how existing families of MDPs are special cases of G-MDP, and therefore map to equivalent DSO problems under (\ref{GMDP}). Only the LMDP family is not a strict special case of G-MDP and will be handled separately. The MDP families we consider are summarized in the following table:
\begin{center}
\begin{tabular}{ l l l}
 symbol & MDP type & step cost \\
 \hline
 G & general & $\ell(x,a,\theta)$ \\
 S & standard, stochastic policy & $r(x,a)$ \\
 D & standard, deterministic policy & $r(x,\mu(x,\theta))$ \\ 
 H & maximum entropy & $r(x,a) + H[\pi(\cdot|x,\theta)]$ \\
 R & proximal regularization & $r(x,a) + D_\textrm{KL}[\pi_\textrm{old} 
 (\cdot|x) \parallel \pi(\cdot|x,\theta) ]$ \\
 L & linearly solvable & $r(x) + D_\textrm{KL}[\textrm{terms outside G-MDP}]$ \\
 \hline
\end{tabular}
\end{center}
We now go over each family and write down its resulting mapping to DSO explicitly.

\textbf{S-MDP} is a standard MDP with stochastic policy. It is the same as G-MDP except the cost $r(x,a)$ does not depend on $\theta$. The mapping is:
\begin{equation}\label{SMDP}
\begin{aligned}
P_S(x'|x,\theta) & \equiv P_G(x'|x,\theta) \\
L_S(x,\theta) & \equiv \int \pi(a|x,\theta) \thinspace r(x,a) \thinspace da
\end{aligned}
\end{equation}

\textbf{D-MDP} is a standard MDP with deterministic policy $a = \mu(x,\theta)$. It is a special case of G-MDP with $\pi(a|x,\theta) = \delta(a - \mu(x,\theta))$ and cost $r(x,a)$ which does not depend on $\theta$. The mapping is:
\begin{equation}\label{DMDP}
\begin{aligned}
P_D(x'|x,\theta) & \equiv p(x'|x,\mu(x,\theta)) \\
L_D(x,\theta) & \equiv r(x,\mu(x,\theta))
\end{aligned}
\end{equation}

\textbf{H-MDP} is an S-MDP which adds the entropy $H\left[\pi(\cdot|x,\theta)\right]$ of the policy to the standard cost $r(x,a)$ so as to encourage exploration. Note that entropy depends on the policy itself and not on the random action sampled from the policy. This type of augmented cost is used in several RL frameworks including maximum entropy \cite{ziebart2010}, soft $Q$-learning \cite{haarnoja2017}, soft actor-critic \cite{haarnoja2018}; also related to the LMDP framework \cite{todorov2006}. The mapping is:
\begin{equation}\label{HMDP}
\begin{aligned}
P_H(x'|x,\theta) & \equiv P_G(x'|x,\theta) \\
L_H(x,\theta) & \equiv \int \pi(a|x,\theta) r(x,a) da + H[\pi(\cdot|x,\theta)]
\end{aligned}
\end{equation}

\textbf{R-MDP} is an S-MDP with extra cost added in each major iteration of policy optimization, aiming to keep the new policy $\pi(\cdot|x,\theta)$ close to the policy $\pi_\textrm{old}(\cdot|x)$ computed in the previous iteration. It is usually KL divergence, but other measures of divergence can be used. Methods that rely on such proximal regularization include TRPO \cite{schulman2015} and MPO \cite{abdolmaleki2018}; a more general formulation is developed in \cite{geist2019}. The mapping is:
\begin{equation}\label{RMDP}
\begin{aligned}
P_R(x'|x,\theta) & \equiv P_G(x'|x,\theta) \\
L_R(x,\theta) & \equiv \int \pi(a|x,\theta) r(x,a) da + D_\textrm{KL}\left[\pi_\textrm{old}(\cdot|x) \parallel \pi(\cdot|x,\theta)\right]
\end{aligned}
\end{equation}

\textbf{L-MDP} is the only case that is not obtained from G-MDP in a straightforward way. It is also the most similar to DSO: it defines actions as state transition probabilities, and penalizes their KL divergence from some baseline (uncontrolled) Markov chain $\bar{p}(x'|x)$. L-MDP can be mapped directly to DSO as:
\begin{equation}\label{LMDP}
\begin{aligned}
P_L(x'|x,\theta) & : \textrm{given} \\
L_L(x,\theta) & \equiv r(x) + D_\textrm{KL}\left[P_L(\cdot|x,\theta) \parallel \bar{p}(\cdot|x)\right]
\end{aligned}
\end{equation}
This mapping is obvious, but we can also derive it using the above general procedure as follows. Define an unusual G-MDP where the next state equals the action, so $n_x = n_a$ and $\pi(a|x,\theta)$ is the same as $P(x'|x,\theta)$. The transition probability and step cost of this G-MDP are:
\begin{equation}
\begin{aligned}
p(x'|x,a) & = \delta(x' - a) \\
\ell(x, a, \theta) & = r(x) + D_\textrm{KL}\left[\pi(\cdot|x,\theta) \parallel \bar{p}(\cdot|x)\right]
\end{aligned}
\end{equation}
Note that here $\ell(x,a,\theta)$ does not depend on $a$. One can now verify that the mapping (\ref{GMDP}) applied to this G-MDP yields the same DSO as the direct mapping above.

\subsection{Pairs of equivalent MDPs that map to the same DSO}

Since we expressed policy optimization in multiple MDP families as DSO problems, an interesting question arises: are some of the resulting DSO problems equivalent, in which case the corresponding MDPs are also equivalent? We recently demonstrated such equivalence \cite{todorov2025} between standard MDPs with stochastic and deterministic policies (corresponding to S-MDP and D-MDP here) in a self-contained way, without reference to DSO. Here we have a simpler procedure at our disposal: show that two MDPs map to the same DSO. This adds to our earlier results from \cite{todorov2025} a newfound equivalence between L-MDP and D-MDP.

The relationship among H-MDP, R-MDP and L-MDP in the present context is clear from their definitions. They differ by costs that do not depend on the actions and are already in the form $L(x,\theta)$. Therefore in this section we will focus on equivalence relations involving S-MDP, D-MDP and L-MDP (as our preferred instantiation of cross-entropy costs.)

\subsubsection*{L-MDP : D-MDP equivalence}

An equivalent pair of an L-MDP and a D-MDP can be constructed by imposing certain restrictions on both sides. In this construction we are free to choose the transition probability $p(x'|x,a)$ and deterministic policy $a = \mu(x,\theta)$ in the D-MDP, and the baseline dynamics $\bar{p}(x'|x)$ and state cost $r_L(x)$ in the L-MDP. Then we set the D-MDP cost to match the L-MDP cost as follows:
\begin{equation}
r(x,a) \equiv r_L(x) + D_\textrm{KL}\left( p(\cdot|x, a) \parallel \bar{p}(\cdot, x) \right)
\end{equation}
Applying the mappings from L-MDP (\ref{LMDP}) and D-MDP (\ref{DMDP}) to DSO, we see that the two are now expressed as the same DSO problem:
\begin{equation}
\begin{aligned}
L_L(x,\theta) & = L_D(x,\theta) = r(x,\mu(x,\theta)) \\
P_L(x'|x,\theta) & = P_D(x'|x,\theta) = p(x'|x,\mu(x,\theta))
\end{aligned}
\end{equation}

\subsubsection*{S-MDP : D-MDP equivalence}

An equivalent pair of an S-MDP and a D-MDP can be constructed by starting with any S-MDP and designing a corresponding D-MDP. The latter has different action space from the S-MDP, yet it maps to the same DSO problem. Let $r(x,a)$, $p(x'|x,a)$ and $\pi(a|x,\theta)$ be the cost, dynamics and stochastic policy of the S-MDP as before. We assume that the stochastic action $a$ is generated from some density in the form $\tilde{\pi}(a|\eta)$ which depends on some minimal set of density parameters (sufficient statistics) denoted $\eta$. These in turn are set by some underlying deterministic policy $\eta = \mu(x,\theta)$. Such a minimal set always exists. Depending on the nature of the density, $\eta$ could be just the mean (in which case $\eta$ and $a$ live in the same space), or the mean and the (state- or parameter-dependent) covariance, or another set of quantities. If there is no deterministic bottleneck in how $\pi(a|x,\theta)$ generates random actions, the default $\eta = (x,\theta)$ is always sufficient. The S-MDP policy can then be written as:
\begin{equation}
\pi(a|x,\theta) = \tilde{\pi}(a | \mu(x,\theta))
\end{equation}

\noindent The corresponding D-MDP has action $\eta$, policy $\eta = \mu(x,\theta)$, and cost and dynamics constructed as:
\begin{equation}
\begin{aligned}
r_D(x,\eta) & \equiv \int \tilde{\pi}(a | \eta)  r(x,a) da \\
p_D(x'|x,\eta) & \equiv \int \tilde{\pi}(a | \eta) p(x'|x,a) da 
\end{aligned}
\end{equation}
We can now verify that the S-MDP we started with, and the D-MDP we constructed, map to the same DSO problem under (\ref{GMDP}). For the costs we have:
\begin{equation}
L_\textrm{D}(x, \theta) = r_D(x,\mu(x,\theta)) = \int \tilde{\pi}(a | \mu(x,\theta)) r(x,a) da = \int \pi(a | x,\theta) r(x,a) da = L_S(x, \theta)
\end{equation}
and similarly for the dynamics:
\begin{equation}
P_D(x'|x, \theta) = p_D(x,\mu(x,\theta)) = \int \tilde{\pi}(a | \mu(x,\theta)) p(x'|x,a) da = \int \pi(a | x,\theta) p(x'|x,a) da = P_S(x'|x, \theta)
\end{equation}
In \cite{todorov2025} we studied a special case of the above equivalence, in a widely-used family of MDPs involving quadratic action cost and Gaussian action noise (while the dynamics and state costs remained general.) In that case the integrals were evaluated in closed form, making the construction more concrete.

Can we combine our two equivalence results in a transitive way, and obtain equivalence between S-MDPs and L-MDPs? This does not seem possible in general. Recall that the directionality is:
\begin{equation}
\textrm{S-MDP} \longrightarrow \textrm{D-MDP} \longleftrightarrow \textrm{L-MDP}
\end{equation}
The circular nature of the D-MDP $\longleftrightarrow$ L-MDP construction is what blocks transitivity.

\section{Computing the DSO gradient}

We now state the DSO gradient results for all definitions of $J(\theta)$, followed by proofs in the next section.

\begin{theorem}[Episodic cost] The gradient of the function $J(\theta)$ defined in (\ref{EPIperf}) is:
\begin{equation}\label{EPItheorem}
\nabla_\theta J(\theta) = \mathbb{E}_{x \sim \rho(\cdot, \theta)} \left[ \nabla_\theta L(x, \theta) + \gamma \int \nabla_\theta P(x'|x, \theta) \thinspace V(x',\theta) \thinspace dx' \right]
\end{equation}
where $V(x,\theta)$ is the function defined in (\ref{EPI_V}) and $\rho(x,\theta)$ is the density defined in (\ref{rho}).
\end{theorem}

\begin{theorem}[Average cost] The gradient of the function $J(\theta)$ defined in (\ref{AVperf}) is:
\begin{equation}\label{AVtheorem}
\nabla_\theta J(\theta) = \mathbb{E}_{x \sim d(\cdot, \theta)} \left[ \nabla_\theta L(x, \theta) +
\int \nabla_\theta P(x'|x, \theta) \thinspace V(x',\theta) \thinspace dx' \right]
\end{equation}
where $V(x,\theta)$ is the unique solution to (\ref{AVbel}) and $d(x,\theta)$ is the unique solution to (\ref{d}).
\end{theorem}

\noindent \textbf{Corollary 1 (Expectation form)} Using the identity $\nabla P = P \thickspace \nabla \ln P$, the results from Theorems 1-2 can be expressed as expectations. We combine the episodic and average cases into a single expression, since they only differ by the desnsity ($\rho$ or $d$) and the value of $\gamma$ -- which is $1$ in average and first-exit settings.
\begin{equation}\label{ExpForm}
\nabla_\theta J(\theta) = \mathbb{E}_{x \sim \rho \textrm{ or } d} \left[ \nabla_\theta L(x, \theta) + \gamma \mathbb{E}_{x' \sim P(\cdot|x,\theta)} \left[ \nabla_\theta \ln P(x'|x,\theta) \thinspace V(x',\theta)\right] \right]
\end{equation}

\noindent \textbf{Corollary 2 (Bottleneck form)} If $\mu : \mathcal{R}^{n_x} \times \mathcal{R}^{n_\theta} \mapsto \mathcal{R}^{n_\eta}$ is such that $L,P$ are in the form:
\begin{equation}
L(x,\theta) = \tilde{L}(x,\mu(x,\theta)), \quad P(x'|x,\theta) = \tilde{P}(x'|x,\mu(x,\theta))
\end{equation}
from some $\tilde{L},\tilde{P}$, then the results from Theorems 1-2 can be expressed as:
\begin{equation}\label{BottleneckForm}
\quad \nabla_\theta J(\theta) = \mathbb{E}_{x \sim \rho \textrm{ or } d} \left[ \nabla_\theta \mu(x,\theta) \left.\left( \nabla_\eta \tilde{L}(x, \eta) + \gamma \int \nabla_\eta \tilde{P}(x'|x, \eta) \thinspace V(x',\theta) \thinspace dx' \right)\right|_{\eta=\mu(x,\theta)} \right]
\end{equation}
This is trivially achieved with $\mu(x,\theta) = \theta, \tilde{L} = L, \tilde{P} = P$. The interesting case is when the parameters act through a ``bottleneck'' $n_\eta < n_\theta$, which arises for example when the DSO is constructed from a D-MDP.

\begin{theorem}[Time-varying cost] The gradient of the function $J(\theta)$ defined in (\ref{TIMEperf}) is:
\begin{equation}\label{TIMEtheorem}
\nabla_\theta J(\theta) = \mathbb{E}_{x_0 \sim P_0(\cdot),\thinspace x_{t+1} \sim P_{t}(\cdot|x_t, \theta)} \left[ \textsum_{t=0}^{T} \nabla_\theta L_t(x_t, \theta) + \textsum_{t=0}^{T-1} \nabla_\theta \ln P_{t}(x_{t+1}|x_t, \theta) \thinspace  V_{t+1}(x_{t+1},\theta) \right]
\end{equation}
where $V_t(x,\theta)$ is the function defined in (\ref{TIME_V}).
\end{theorem}

\subsection{Proofs of the theorems}

The proofs here are simplified versions of the corresponding MDP policy gradient theorems. The procedure in each case is to differentiate the Bellman equation, average it over the relevant density, and obtain the DSO gradient after simplification and grouping of terms.

\subsubsection*{Proof of Theorem 1}

Differentiating $J(\theta)$ from (\ref{EPIperf}) yields:
\begin{equation}\label{EPIJdif}
\nabla_\theta J(\theta) = \mathbb{E}_{x \sim P_0(\cdot)} \left[ \nabla_\theta V(x, \theta) \right]
\end{equation}

\noindent Now we differentiate the Bellman equation (\ref{EPIbel}):
\begin{equation}
\nabla_\theta V(x, \theta) = \nabla_\theta L(x, \theta) + \gamma
\int \nabla_\theta P(x'|x, \theta) V(x', \theta) dx' + \gamma \int P(x'|x, \theta) \nabla_\theta V(x', \theta) dx'
\end{equation}
This relates $\nabla V$ at the current state $x$ and at the next state $x'$, so it is a recursion but the quantities are not time-varying and the recursion extends to infinity. To avoid long subscripts in the expectations, define $P_X(X)$ as the probability of a trajectory $X = (x_0, x_1, \cdots)$ under the Markov chain.

Now we replace $\nabla P$ with $P \thinspace \nabla \ln P$ so as to switch to expectation form; unfold the recursion to infinity; multiply both sides by $P_0(x)$; and integrate over $x$. This yields:
\begin{equation}\label{mess}
\begin{aligned}
\mathbb{E}_{x \sim P_0(\cdot)} \left[ \nabla_\theta V(x, \theta) \right] & = \mathbb{E}_{X \sim P_X(\cdot,\theta)} \left[ \textsum_{t=0}^{\infty} \left(\gamma^t \nabla_\theta L(x_t, \theta) + \gamma^{t+1} \nabla_\theta \ln P(x_{t+1}|x_t,\theta) V(x_{t+1},\theta)\right)\right] \\
& + \mathbb{E}_{X \sim P_X(\cdot,\theta)} \left[\lim_{t\rightarrow\infty} \gamma^t \nabla_\theta V(x_t, \theta) \right]
\end{aligned}
\end{equation}
On the left we have the gradient from (\ref{EPIJdif}). The limit term on the right is $0$ because of the exponential discounting. In a first-exit formulation where we have a set $\mathcal{T}$ of terminal states that are absorbing and incur zero cost, $\nabla_\theta V(x,\theta) = 0$ for $x \in \mathcal{T}$ and so that term is again zero. As for the first term on the right, we can simplify notation using the quantity:
\begin{equation}\label{Mdef}
K(x,x',\theta) \equiv \nabla_\theta L(x, \theta) + \gamma \nabla_\theta \ln P(x'|x,\theta) V(x',\theta)
\end{equation}
Equation (\ref{mess}) now becomes:
\begin{equation}
\nabla_\theta J(\theta) = \mathbb{E}_{X \sim P_X(\cdot,\theta)} \left[ \textsum_{t=0}^{\infty} \gamma^t K(x_t,x_{t+1},\theta) \right] 
= \mathbb{E}_{x \sim \rho(\cdot,\theta), \thinspace x' \sim P(\cdot|x,\theta)} \left[ K(x,x',\theta) \right]
\end{equation}
The last step follows from the definition of the discounted/visitation density $\rho(x,\theta)$ in (\ref{rho}). Substituting the definition of $K$ now yields the result in the theorem. $\square$

\subsubsection*{Proof of Theorem 2}

Differentiating the Bellman equation (\ref{AVbel}) yields:
\begin{equation}\label{AVbeldif}
\nabla_\theta J(\theta) + \nabla_\theta V(x, \theta) = \nabla_\theta L(x, \theta) + 
\int \nabla_\theta P(x'|x, \theta) V(x', \theta) dx' + \int P(x'|x, \theta) \nabla_\theta V(x', \theta) dx'   
\end{equation}
Now we multiply both sides by $d(x,\theta)$ and integrate over $x$. The Left hand side of (\ref{AVbeldif}) becomes:
\begin{equation}
\textrm{LHS:} \quad \nabla_\theta J(\theta) + \int d(x,\theta) \nabla_\theta V(x, \theta) dx
\end{equation}
The Right hand side of (\ref{AVbeldif}) becomes:
\begin{equation}
\textrm{RHS:} \quad \mathbb{E}_{x \sim d(\cdot, \theta)} \left[\nabla_\theta L(x, \theta) + 
\int \nabla_\theta P(x'|x, \theta) V(x', \theta) dx' \right] + \iint d(x,\theta) P(x'|x, \theta) \nabla_\theta V(x', \theta) dx' dx
\end{equation}
Since $d(x)$ is stationary under $P(x'|x)$, for any function $g(x)$ we have the generic identity:
\begin{equation}
\int d(x) g(x) dx = \iint d(x) P(x'|x) g(x') dx' dx
\end{equation}
Applying this identity with $g = \nabla V$ shows that the last terms in LHS and RHS are equal, so they cancel each other, and we are left with:
\begin{equation}
\nabla_\theta J(\theta) = \mathbb{E}_{x \sim d(\cdot, \theta)} \left[\nabla_\theta L(x, \theta) + 
\int \nabla_\theta P(x'|x, \theta) V(x', \theta) dx' \right]
\end{equation}
This is the same result as in the theorem. $\square$

\subsubsection*{Proof of Theorem 3}

Differentiating $J(\theta)$ from (\ref{TIMEperf}) yields:
\begin{equation}\label{TIMEJdif}
\nabla_\theta J(\theta) = \mathbb{E}_{x \sim P_0(\cdot)} \left[ \nabla_\theta V_0(x, \theta) \right]
\end{equation}

\noindent Now we differentiate the Bellman equation (\ref{TIMEbel}):
\begin{equation}\label{TIMEbeldif}
\nabla_\theta V_t(x, \theta) = \nabla_\theta L_t(x, \theta) + \int \nabla_\theta P_t(x'|x, \theta) V_{t+1}(x', \theta) dx' + \int P_t(x'|x, \theta) \nabla_\theta V_{t+1}(x', \theta) dx'
\end{equation}
This is a recursion over time for $\nabla V_t$. But unlike the episodic case where it could continue indefinitely, here it ends at the final time $T$ with $\nabla_\theta V_T(x, \theta) = \nabla_\theta L_T(x, \theta)$. We start from $t=0$, expand (\ref{TIMEbeldif}) recursively all the way to $t = T$, and replace $\nabla P$ with $P \thinspace \nabla \ln P$ so as to switch to expectation form. This yields:
\begin{equation}
\nabla_\theta V_0(x, \theta) = \mathbb{E}_{x_0 = x,\thinspace x_{t+1} \sim P_{t}(\cdot|x_t, \theta)} \left[ \textsum_{t=0}^{T} \nabla_\theta L_t(x_t, \theta) + \textsum_{t=0}^{T-1} \nabla_\theta \ln P_{t}(x_{t+1}|x_t, \theta) \thinspace  V_{t+1}(x_{t+1},\theta) \right]
\end{equation}
Multiply both sides by $P_0(x)$ and integrate over $x$. Applying (\ref{TIMEJdif}) yields the result in the theorem. $\square$

\subsection{MDP policy gradients as DSO gradients}

We have already shown in Section 2 that existing families of MDPs can be mapped to DSO, such that the expected policy performance $\mathcal{J}(\theta)$ for each MDP equals the corresponding DSO objective $J(\theta)$. Therefore the policy gradient $\nabla_\theta \mathcal{J}(\theta)$ in these MDPs equals the DSO gradient $\nabla_\theta J(\theta)$ as given by Theorems 1-3.

Here we make this correspondence more explicit, as a sanity check but also for added clarity. In each case we show through direct derivation that the known MDP policy gradient equals the DSO gradient. The equality of the MDP and DSO value functions as well as transition probabilities was already established in Section 2; we now use it as a known fact, and derive $\nabla_\theta \mathcal{J}(\theta) =\nabla_\theta J(\theta)$ from it.

\subsubsection*{S-MDP policy gradient}

The policy gradient for standard MDPs with stochastic policies was given in \cite{sutton1999} (Theorem 1 in the reference.) That result is:
\begin{equation}\label{SPG}
\nabla_\theta \mathcal{J}(\theta) =  \mathbb{E}_{x \sim \rho(\cdot, \theta)} \left[ \int \nabla_\theta \pi(a|x, \theta) Q(x, a, \theta) da \right]
\end{equation}
where the $Q$ function is:
\begin{equation}\label{Qfunc}
Q(x,a,\theta) = r(x,a) + \gamma \int p(x'|x,a) v(x',\theta) dx'
\end{equation}
The term inside the expectation in (\ref{SPG}) can be expanded as:
\begin{equation}
\begin{aligned}
\int \nabla_\theta \pi(a|x, \theta) Q(x, a, \theta) da & = 
\int \nabla_\theta \pi(a|x, \theta) r(x,a) da + \gamma \iint \nabla_\theta \pi(a|x, \theta)  p(x'|x, a) v(x',\theta) dx' da \\
& = \underbrace{\nabla_\theta \int \pi(a|x, \theta) r(x,a) da}_{\nabla_\theta L_S(x'|x,\theta)} + \gamma \int \underbrace{\left( \nabla_\theta \int  \pi(a|x, \theta) p(x'|x, a) da \right)}_{\nabla_\theta P_S(x'|x,\theta)} v(x',\theta) dx'
\end{aligned}
\end{equation}
where we have recognized the definitions (\ref{SMDP}). The last expression coincides with the result in Theorem 1. Thus the S-MDP policy gradient equals the corresponding DSO gradient.

\subsubsection*{D-MDP policy gradient}

The policy gradient for standard MDPs with deterministic policies was given in \cite{silver2014} (Theorem 1 in the reference.) It involves differentiation over actions, so here $a$ must be continuous. That result is:
\begin{equation}\label{DPG}
\nabla_\theta \mathcal{J}(\theta) = \mathbb{E}_{x \sim \rho(\cdot, \theta)} \left[ \nabla_\theta \mu(x, \theta) \left. \nabla_a Q(x,a,\theta) \right|_{a = \mu(x, \theta)} \right]
\end{equation}
The $Q$ function is as in (\ref{Qfunc}). Substituting (\ref{Qfunc}) in (\ref{DPG}), the term inside the expectation in (\ref{DPG}) becomes:
\begin{equation}
\begin{aligned}
\nabla_\theta \mu(x, \theta) \left. \nabla_a Q(x,a,\theta) \right|_{a = \mu} & = 
\nabla_\theta \mu(x,\theta) \left. \nabla_a r(x,a) \right|_{a=\mu} + 
\gamma \int \nabla_\theta \mu(x,\theta) \left. \nabla_a p(x'|x,a)  \right|_{a = \mu} v(x',\theta) dx' \\
& = \underbrace{\nabla_\theta r(x, \mu(x, \theta))}_{\nabla_\theta L_D(x, \theta)} + \gamma \int \underbrace{\nabla_\theta p(x'|x,\mu(x, \theta))}_{\nabla_\theta P_D(x'|x,\theta)} v(x',\theta) dx'
\end{aligned}
\end{equation}
where we have recognized the definitions (\ref{DMDP}). The last expression coincides with the result in Theorem 1. Thus the D-MDP policy gradient equals the corresponding DSO gradient.

\subsubsection*{L-MDP policy gradient}

The policy gradient for Linearly-solvable MDPs was given in \cite{todorov2010} (Theorem 1 in the reference.) It was derived in the average cost setting. That result is:
\begin{equation}\label{LMDPequiv}
\nabla_\theta \mathcal{J}(\theta) =  \mathbb{E}_{x \sim d(\cdot, \theta)} \left[ \int \nabla_\theta P(x'|x,\theta) \left( \ln \frac{P(x'|x,\theta)}{\bar{p}(x'|x)} + v(x',\theta)\right) dx' \right]
\end{equation}
In order to show that this coincides with the result in Theorem 2, it is sufficient to show that the term:
\begin{equation}
\int \nabla_\theta P(x'|x,\theta) \ln \frac{P(x'|x,\theta)}{\bar{p}(x'|x)} dx' 
\end{equation}
appearing in (\ref{LMDPequiv}) is equal to $\nabla_\theta L_L(x,\theta)$. Differentiating the definition of $L_L$ from (\ref{LMDP}) yields:
\begin{equation}
\begin{aligned}
\nabla_\theta L_L(x,\theta) & = 
\nabla_\theta \int P(x'|x,\theta) \ln \frac{P(x'|x,\theta)}{\bar{p}(x'|x)} dx' \\
& = \int \nabla_\theta P(x'|x,\theta) \ln \frac{P(x'|x,\theta)}{\bar{p}(x'|x)} dx' + 
\int P(x'|x,\theta) \nabla_\theta \ln P(x'|x,\theta) dx'
\end{aligned}
\end{equation}
The second integral can be written as $\nabla_\theta \int P(x'|x,\theta) dx' = \nabla_\theta 1 = 0$. Thus the L-MDP policy gradient equals the corresponding DSO gradient.

\section{Optimizing the DSO objective}

Once we estimate the gradient of the DSO objective using Theorems 1-3, we can feed it into a stochastic gradient descent optimizer such as ADAM. In this section we improve upon the baseline approach by adapting more refined methods from the MDP literature.

\subsection{Variance reduction}

Starting with the REINFORCE algorithm \cite{williams1992}, stochastic approximations to MDP policy gradients have suffered from high variance when used in their pure form. Variance can be reduced with help from a state value function approximation $\hat{V}(x,\omega)$, fitted by minimizing Bellman errors with respect to $\omega$ for fixed $\theta$. This topic has been extensively studied in MDPs and the same methods apply here as well. Fitting is a supervised learning problem for the approximation parameters $\omega$. The supervised learning loss can penalize deviations from empirical returns, or Bellman errors, or ``interpolate'' between the two using Temporal Difference methods \cite{sutton2018}. The resulting $\hat{V}(x,\omega)$ can then be used for advantage estimation \cite{schulman2016}.

In DSO we do not have advantages (since they involve actions) however we can apply the same machinery in state space, as follow. Before replacing expectations with samples, subtract a baseline term that leaves the expectations unchanged -- so that the stochastic approximation remains unbiased. A suitable baseline here will be obtained by combining the value function approximation with a model of deterministic dynamics: $x' = f(x,\theta)$. The latter should be close (ideally equal) to the mean transition of the Markov chain.

The construction is based on the identity:
\begin{equation}
\int \nabla_\theta P(x'|x,\theta) V(x',\theta) dx'
= \int \nabla_\theta P(x'|x,\theta) \left(V(x',\theta) - \hat{V}(f(x,\theta))\right) dx'
\end{equation}
This identity holds because $\hat{V}(f(x,\theta))$ does not depend on $x'$. Expanding the right hand side, the second term is $\hat{V} \nabla_\theta \int P(x'|x,\theta) dx'= \hat{V} \nabla_\theta 1 = 0$.
Switching to expectations on the right, we now have:
\begin{equation}
\int \nabla_\theta P(x'|x,\theta) V(x',\theta) dx' = 
\mathbb{E}_{x' \sim P(\cdot|x,\theta)}\left[\nabla_\theta \ln P(x'|x,\theta) \thinspace \left(V(x',\theta) - \hat{V}(f(x,\theta))\right)\right]
\end{equation}
In a stochastic approximation scheme using trajectory rollouts, the integral term on the left (which is the same term appearing in Theorems 1-3) is then replaced with the sample-based term:
\begin{equation}
\nabla_\theta \ln P(x_{t+1}|x_t,\theta) \left( R_{t+1} - \hat{V}(f(x_t,\theta))\right)
\end{equation}
Here $R_t$ denotes the empirical cumulative cost starting from step $t$ in each trajectory rollout.

Putting everything together, we have an algorithm for estimating the DSO gradient from a batch of rollouts. The function approximation $\hat{V}$ is fitted to data up to the previous batch, but not the current batch. Rollout termination can occur at predefined times, terminal states (which can also be used to detect simulation instabilities) or randomly with probability $(1-\gamma)$ at each step. This sampling scheme corresponds to the visitation-density interpretation of $\rho$ in the discounted setting. Thus we treat termination as a decision made by the rollout itself. In time-varying settings, we use time-varying $L_t$ and $P_t$ and the rest remains the same. We can even use rollout-specific $L_t^n$ and $P_t^n$.

\begin{algorithm}
  \caption{DSO Gradient Estimation}
  \begin{algorithmic}
    \For {$n = 0 : N-1$} \Comment{batch of trajectory rollouts}
      \State $x_0^n \sim P_0(\cdot)$ 
      \While {\textrm{rollout not ended}} \Comment{$T^n$ is the step at which the rollout ends}
        \State $x_{t+1}^n \sim P\left(\cdot|x_t^n,\theta\right)$
      \EndWhile
      \State $R_{T^n}^n \gets L\left(x_{T^n}^n,\theta\right)$ \Comment {$R_t^n$ is the cumulative cost starting from $t$}
      \State $G_{T^n}^n \gets \nabla_\theta L\left(x_{T^n}^n,\theta\right)$ \Comment{$G_t^n$ is the cumulative gradient starting from $t$}
      \For {$t = T^n - 1 : 0$}
        \State $G_t^n \gets \gamma G_{t+1}^n + \nabla_\theta L(x_t^n,\theta) + \gamma \nabla_\theta \ln P(x_{t+1}^n|x_t^n,\theta) \left(R_{t+1}^n - \hat{V}(f(x_t^n,\theta))\right)$
        \State $R_t^n \gets \gamma R_{t+1}^n + L(x_t^n,\theta)$
      \EndFor
    \EndFor
    \State $\tilde{\nabla}_\theta J \gets \tfrac{1}{N} \sum_{n=0}^{N-1} G_0^n$  \Comment{return the mean of the rollout gradients}
  \end{algorithmic}
\end{algorithm}

\subsection{Surrogate objective and DSO Chain Iteration}

We now seek a surrogate objective $S$ whose gradient equals the gradient of the DSO objective $J$, and can be obtained by direct differentiation. Note that the $J$ itself does not satisfy this requirement. Indeed Theorems 1-3 involve gradients of the defining quantities $P,L$ but no gradients of the derived quantities $V,d,\rho$. If we were to apply $\nabla_\theta$ to an expression involving all of the above quantities, we would end up with undesirable gradients of $V,d,\rho$. Thus a more subtle approach is needed.

The above considerations motivate the following surrogate objective in the episodic setting:
\begin{equation}\label{Surrogate}
S(\theta,\alpha) \equiv \mathbb{E}_{x \sim \rho(\cdot, \theta)} \left[ L\left(x,\theta + \alpha\right) + \gamma \int P\left(x'|x,\theta+\alpha\right) V(x',\theta) dx' \right]
\end{equation}
The average setting is the same, but using $d(x,\theta)$ instead of $\rho(x,\theta)$ and setting $\gamma = 1$. The vector $\alpha$ has the semantics of a perturbation to $\theta$. However this perturbation is only applied to $P,L$ and not to $V,d,\rho$.

\begin{lemma}[Surrogate gradient equals DSO gradient] Let $S$ be as defined in (\ref{Surrogate}) and $J$ be as defined in (\ref{EPIperf}). Then for all $\theta$,
\begin{equation}\label{GradMatch}
\nabla_\alpha \left. S(\theta, \alpha) \right|_{\alpha = 0} = \nabla_\theta J(\theta)
\end{equation}
This also holds when $J$ is as defined in (\ref{AVperf}), $\rho$ is replaced with $d$ in (\ref{Surrogate}) and $\gamma = 1$.
\end{lemma}
\textbf{Proof.} 
Differentiating (\ref{Surrogate}) with respect to $\alpha$ and setting $\alpha = 0$ yields the same expression as the DSO gradient result in Theorem 1 and 2 respectively. $\square$

Note that the surrogate $S$ is not intended to approximate the DSO objective $J$ globally, and indeed it would be a very poor approximation. For concreteness, suppose the costs $L$ are non-negative. Then $S(\theta,0) \gg J(\theta)$. This is because $S$ adds up the costs \textit{and} the values, while the values already incorporate all future costs; so $S$ massively over-counts the costs. Nevertheless, given that the gradients match, $S$ can be used to analyze local changes in $J$. Assuming both functions are smooth, (\ref{GradMatch}) implies that for all $\theta$,
\begin{equation}
J(\theta + \alpha) - J(\theta) = S(\theta,\alpha) - S(\theta,0) + o\left(\|\alpha\|\right)
\end{equation}

\noindent This surrogate also enables a DSO analog to policy iteration. Instead of merely using the gradient of $S$ to obtain the gradient of $J$, we can optimize $S(\theta, \alpha)$ over $\alpha$ for fixed $\theta$. This yields DSO Chain Iteration:
\begin{equation}
\begin{aligned}
\alpha_k^* & = \arg \min_\alpha S(\theta_k, \alpha) \\
\theta_{k+1} & = \theta_{k} + \alpha_k^*
\end{aligned}
\end{equation}
Reducing the stepsize as $\theta_{k+1} = \theta_{k} + \kappa \alpha_k^*$ would then be similar to Conservative Policy Iteration \cite{kakade2002}. In the MDP literature, expressing the policy gradient as the derivative of the Bellman policy iteration operator has proven useful in theoretical analysis \cite{bhandari2024}, and is related to our surrogate objective.

\subsection{Proximal optimization}

Proximal methods in MDPs aim to limit the step taken by gradient descent, so that the visitation/stationary density does not change too much. This is usually done by penalizing KL divergence between the new and old policies, as in the R-MDP discussed earlier. We can implement proximal optimization here by simply optimizing a DSO objective that came from an R-MDP. We can also do it directly at the DSO level, by adding $D_\textrm{KL}\left(P(\cdot|x,\theta_\textrm{old}) \parallel P(\cdot|x,\theta)\right)$ to the DSO cost $L(x,\theta)$.

Another aspect of proximal optimization, introduced by TRPO \cite{schulman2015} and then inherited by PPO \cite{schulman2017} and related methods, is to replace the terms $(\nabla \ln \pi)$ in the policy gradient with a likelihood ratio -- which was then clipped in PPO. We can do the same here using our surrogate $S$. To this end we express $S$ as expectation, and apply the importance sampling transformation to make sure all expectations are taken under the original chain $P(x'|x,\theta)$ and not the perturbed chain $P(x'|x,\theta + \alpha)$. The result is:
\begin{equation}
\begin{aligned}
S(\theta,\alpha) & = \mathbb{E}_{x \sim \rho(\cdot, \theta)} \left[ L\left(x,\theta + \alpha\right) + \gamma \mathbb{E}_{x' \sim P\left(\cdot|x,\theta+\alpha\right)} \left[ V(x',\theta) \right] \right] \\
& = \mathbb{E}_{x \sim \rho(\cdot, \theta)} \left[ L\left(x,\theta + \alpha\right) + \gamma \mathbb{E}_{x' \sim P(\cdot|x,\theta)} \left[ \frac{P\left(x'|x,\theta+\alpha\right)}{P(x'|x,\theta)} V(x',\theta) \right] \right]
\end{aligned}
\end{equation}
Next we replace the true $S$ with its sampling approximation $\tilde{S}$, constructed in the same way as the DSO gradient estimate in Algorithm 1. Define the shortcut notation:
\begin{equation}
\begin{aligned}
\hat{L}_t^n(\theta,\alpha) & \equiv L(x_t^n,\theta+\alpha) \\
\hat{P}_{t,t+1}^n(\theta,\alpha) & \equiv \frac{P\left(x_{t+1}^n|x_t^n,\theta+\alpha\right)}{P(x_{t+1}^n|x_t^n,\theta)} \\
\hat{R}_{t,t+1}^n(\theta) & \equiv R_{t+1}^n - \hat{V}(f(x_t^n,\theta))
\end{aligned}
\end{equation}
Then the sampling approximation to the surrogate is:
\begin{equation}
\tilde{S}(\theta,\alpha) \equiv \frac{1}{N}\sum_{n=0}^{N-1} \sum_{t=0}^{T^n} \gamma^t \left( \hat{L}_t^n( \theta, \alpha) + \gamma \thinspace \hat{P}_{t,t+1}^n(\theta,\alpha) \hat{R}_{t,t+1}^n(\theta)\right)
\end{equation}
A PPO-like procedure would clip the terms $ \hat{P}_{t,t+1}^n$; in the present context this may be called Proximal Chain Optimization (PCO). It is now easy to verify that the sample-based estimate of the DSO gradient given by Algorithm 1 matches the analytical gradient of the sample-based estimate of the surrogate:
\begin{equation}
\tilde{\nabla}_\theta J(\theta) = \left. \nabla_\alpha \tilde{S}(\theta, \alpha) \right|_{\alpha=0}
\end{equation}
Writing the latter explicitly, we have:
\begin{equation}\label{SampleGrad}
\nabla_\alpha \tilde{S}(\theta,\alpha) = \frac{1}{N}\sum_{n=0}^{N-1} \sum_{t=0}^{T^n} \gamma^t \left( \nabla_\alpha \hat{L}_t^n( \theta, \alpha) + \gamma \thinspace \nabla_\alpha \hat{P}_{t,t+1}^n(\theta,\alpha) \hat{R}_{t,t+1}^n(\theta)\right)
\end{equation}

\subsection{Natural gradients}

Natural gradients \cite{amari1998,kakade2001,peters2008} are rooted in Differential Geometry. Consider a manifold parameterized by $\theta$ and a scalar function $J(\theta)$ on the manifold. The gradient $\nabla J$ is a vector in the tangent space at fixed $\theta$. The vector of partial derivatives $dJ = (\partial J/\partial \theta_i)$ is a differential form which lives in the dual (or co-tangent) space. Given a metric $M(\theta)$ on the manifold\footnote{The metric $M$, vector $\nabla J$ and co-vector $dJ$ are tensors, which exist independent of a coordinate representation. But here we are working in the coordinates given by $\theta$.}, the mapping from co-tangent to tangent vectors is:
\begin{equation}
\nabla J = M^{-1} dJ
\end{equation}
It is common practice outside Differential Geometry to use $dJ$ and call it $\nabla J$, even though the two are only equal under the Cartesian metric $M = I$. For general metrics, the above correction should be applied. What it does is find the steepest ascent direction, where direction vector lengths are now measured as $v^T M v$.

While $M$ can be any Riemannian metric, it has been argued in the context of supervised learning (and later, policy gradients) that the ``natural'' metric is Fisher information, denoted $F(\theta)$ below. This comes from statistical manifolds. For optimization purposes, a better metric is the Hessian of $J(\theta)$ or a Gauss-Newton approximation to it -- in which case the metric correction yields (Gauss) Newton optimization \cite{martens2020}. However, algorithms that estimate the policy Hessian are very rare; one example is \cite{shen2019}.

In the context of stochastic policy gradients, the expected Fisher information is \cite{kakade2001}:
\begin{equation}\label{FS}
\mathcal{F}_S(\theta) = \mathbb{E}_{x \sim \rho(\cdot, \theta), \thickspace a \sim \pi(\cdot|x, \theta)} \left[ \nabla_\theta \ln \pi(a|x,\theta) \thickspace \nabla_\theta \ln \pi(a|x,\theta)^T \right]
\end{equation}
For deterministic polices, the following analog has been proposed \cite{silver2014}:
\begin{equation}\label{FD}
\mathcal{F}_D(\theta) = \mathbb{E}_{x \sim \rho(\cdot, \theta)} \left[ \nabla_\theta \mu(x,\theta) \thickspace \nabla_\theta \mu(x,\theta)^T \right]
\end{equation}

\noindent We can apply the same idea in DSO:
\begin{equation}\label{DSO_F}
F(\theta) = \mathbb{E}_{x \sim \rho(\cdot, \theta), \thinspace x' \sim P(\cdot|x,\theta)} \left[ \nabla_\theta \ln P(x'|x,\theta) \thickspace \nabla_\theta \ln P(x'|x,\theta)^T \right]
\end{equation}
Since both S-MDP and D-MDP were expressed in DSO (for different $P$ and $L$) the above construction covers both cases. Nevertheless, we can expose additional structure resembling (\ref{FD}) at the DSO level. When the DSO is in the bottleneck form (Corollary 2), we can express (\ref{DSO_F}) as:
\begin{equation}
F(\theta) = \mathbb{E}_{x \sim \rho(\cdot, \theta)} \left[ \nabla_\theta \mu(x,\theta) \left.\left( \mathbb{E}_{x' \sim \tilde{P}(\cdot|x,\eta)} \left[ \nabla_\eta \ln \tilde{P}(x'|x,\eta) \thickspace \nabla_\eta \ln \tilde{P}(x'|x,\eta)^T \right] \right)\right|_{\eta = \mu(x,\theta)} \nabla_\theta \mu(x,\theta)^T \right]
\end{equation}
As a concrete example, consider the Gaussian case $\tilde{P}(x'|x,\eta) = \mathcal{N}(x';\eta,\Sigma(x))$. Then:
\begin{equation}
F(\theta) = \mathbb{E}_{x \sim \rho(\cdot, \theta)} \left[ \nabla_\theta \mu(x,\theta) \Sigma(x)^{-1} \nabla_\theta \mu(x,\theta)^T \right]
\end{equation}
In our recent work \cite{todorov2025} focusing on Gaussian MDP policies, we similarly inserted $\Sigma^{-1}$ in (\ref{FD}).

\subsection{Tangent-space interpretation of the surrogate objective}

The Differential Geometry perspective brought by natural gradients enables an interpretation of the surrogate function $S$ and the perturbation vector $\alpha$, helping understand their relation to the underlying DSO problem. The DSO objective $J(\theta)$ is a scalar function on the manifold parameterized by $\theta$ as before. Now, at a fixed point $\theta_1$ on the manifold, think of $\alpha$ as a vector in the tangent space and $S(\theta_1,\alpha)$ as a scalar function over the tangent space. If we move to another point $\theta_2$, the tangent space changes, and $S(\theta_2, \alpha)$ becomes a scalar function over the new tangent space. In other words $S(\theta,\alpha)$ is a continuous set of functions of the vector argument $\alpha$, while $\theta$ is the index in that set. This is called a function over the tangent bundle.

In this setting, what does the gradient matching condition (\ref{GradMatch}) mean? It means that the directional derivative of $S$ along $\alpha$ in the tangent space equals the rate of change of $J$ along any path on the manifold that passes though the fixed $\theta$ and is tangent to $\alpha$. So the gradient of $S$ yields a direction for gradient descent to move on the manifold.

Recall that we also have DSO Chain Iteration, where we compute an actual step $\alpha^*$ and not just a direction. The earlier update $\theta \leftarrow \theta + \alpha^*$ is suitable for a flat manifold. Ideally, we would move along the geodesic path tangent to $\alpha^*$ by geodesic distance $\|\alpha^*\|$. This mapping from the tangent space to the manifold is called the exponential map. Unfortunately, in order to construct it explicitly, we would have to solve the geodesic equation -- which is a second-order ODE involving derivatives of the metric (namely the Christoffel symbols of the corresponding Levi-Civita connection.) And since the metric here is estimated from samples, differentiating it is not feasible. Still, this interpretation of $S$ and $\alpha$ is useful conceptually.

\subsection{Computing the DSO Hessian}

Even though the Hessians of $S$ and $J$ are not the same, we can use the approximation:
\begin{equation}
\nabla_\theta^2 J(\theta) \approx \left. \nabla_\alpha^2 \tilde{S}(\theta, \alpha) \right|_{\alpha=0}
\end{equation}
The Hessian of the sample-based surrogate is obtained by directly differentiating its gradient (\ref{SampleGrad}):
\begin{equation}
\nabla_\alpha^2 \tilde{S}(\theta,\alpha) = \frac{1}{N}\sum_{n=0}^{N-1} \sum_{t=0}^{T^n} \gamma^t \left( \nabla_\alpha^2 \hat{L}_t^n( \theta, \alpha) + \gamma \thinspace \nabla_\alpha^2 \hat{P}_{t,t+1}^n(\theta,\alpha) \hat{R}_{t,t+1}^n(\theta)\right)
\end{equation}
In DSO Chain Iteration, this Hessian can be used to implement a second-order method for finding $\alpha*$. Note that we are only differentiating terms involving the functions $L$ and $P$ which define the DSO problem, and none of the derived quantities. When $\theta$ is high-dimensional the Hessian matrix is large, but the Fisher matrix $F(\theta)$ is the same size (although the latter does need to be computed explicitly.)

How does this approximation relate to the exact Hessian? While we are not able to provide error bounds, the surrogate has an interesting symmetry which offers some insight. Differentiating the gradient matching condition (\ref{GradMatch}) yields:
\begin{equation}
\nabla_\theta^2 J(\theta) = \nabla_\theta \left( \left.\nabla_\alpha S(\theta, \alpha)\right|_{\alpha=0} \right) = \left. \nabla_\alpha \left( \nabla_\theta S(\theta, \alpha)\right)\right|_{\alpha=0}
\end{equation}
The second equality is non-trivial. It follows from the fact that the Hessian of $J$ is symmetric, and therefore the cross-derivatives of $S$ are symmetric (which is not the case for a generic function of two vector arguments.) We can write this symmetry property more explicitly as:
\begin{equation}
\left. \frac{\partial^2 S(\theta,\alpha)}{\partial \theta_i \partial \alpha_j} \right|_{\alpha = 0} =
\left. \frac{\partial^2 S(\theta,\alpha)}{\partial \theta_j \partial \alpha_i} \right|_{\alpha = 0}
\end{equation}
The implication is that if we somehow obtain a sample-based estimate of $\nabla_\theta S$, we could then differentiate it analytically over $\alpha$, yielding an unbiased estimate of the Hessian of $J$. It would be interesting in future work to look for a value-like function behind $S$, satisfying a Bellman-like equation which can then be used to express $\nabla_\theta S$ in terms of $\nabla_\theta L$ and $\nabla_\theta P$ only.

In the time-varying setting, we can obtain an unbiased Hessian estimate directly. This is because all relevant quantities are given explicitly in terms of $P$ and $L$, and we do not have to deal with infinite-horizon propagation of derivatives. The result is stated below. It can be thought of as second-order REINFORCE adapted to DSO. This Hessian estimate likely has high variance; unless we also learn a (time-varying) value function approximation and apply the earlier variance-reduction procedure.

\begin{lemma}[Unbiased Hessian estimate] In the time-varying setting (\ref{TIMEperf}, \ref{TIMEbel}), let $X = (x_0, x_1, \cdots x_{T})$ denote a path of the Markov chain. Define the cost, probability and log-probability of the path:
\begin{equation}
\begin{aligned}
\mathcal{L}(X,\theta) & \equiv \textsum_{t=0}^T L_t(x_t,\theta) \\
\mathcal{P}(X,\theta) & \equiv P_0(x_0) \textprod_{t=0}^{T-1} P_t(x_{t+1}|x_t,\theta) \\
\mathcal{K}(X,\theta) & \equiv \ln \mathcal{P}(X,\theta) = \ln P_0(x_0) + \textsum_{t=0}^{T-1} \ln P_t(x_{t+1}|x_t,\theta)
\end{aligned}
\end{equation}
Then the performance $J(\theta)$, and its gradient and Hessian with respect to $\theta$ are:
\begin{equation}
\begin{aligned}
J(\theta) & = \mathbb{E}_{X \sim \mathcal{P}(\cdot,\theta)} \left[ \mathcal{L}(X,\theta) \right]  \\
\nabla_\theta J(\theta) & = \mathbb{E}_{X \sim \mathcal{P}(\cdot,\theta)} \left[ \nabla_\theta \mathcal{K} \thinspace \mathcal{L} + \nabla_\theta \mathcal{L}\right] \\
\nabla_\theta^2 J(\theta) & = \mathbb{E}_{X \sim \mathcal{P}(\cdot,\theta)} \left[ \left( \nabla_\theta \mathcal{K} \thinspace \nabla_\theta \mathcal{K}^T + \nabla_\theta^2 \thinspace \mathcal{K}\right) \mathcal{L} + \nabla_\theta \mathcal{K} \thinspace \nabla_\theta \mathcal{L}^T + \nabla_\theta \mathcal{L} \thinspace \nabla_\theta \mathcal{K}^T + \mathcal{K} \thinspace \nabla_\theta^2 \mathcal{L} \right] \\
\end{aligned}
\end{equation}
\end{lemma}
\textbf{Proof.} To reduce clutter, we have dropped the explicit dependence on $(X,\theta)$ in all derivatives. The performance $J(\theta)$ is the expected cost of the path, which coincides with the definition. Writing it as an integral and differentiating, and using the identity $\nabla \mathcal{P} = \mathcal{P} \thinspace \nabla \mathcal{K}$ yields:
\begin{equation}
\nabla_\theta J(\theta) = \nabla_\theta \int \mathcal{P} \thinspace \mathcal{L} \thinspace dX = \int \left(\nabla_\theta \mathcal{P} \thinspace \mathcal{L} + \mathcal{P} \thinspace \nabla_\theta \mathcal{L} \right) dX = \int \mathcal{P} \thinspace \left(\nabla_\theta \mathcal{K} \thinspace \mathcal{L} + \nabla_\theta \mathcal{L} \right) \thinspace dX
\end{equation}
This is the gradient result in the lemma. Now we differentiate the latter integral one more time, use the above identity repeatedly, and group terms. This yields the Hessian result in the lemma. $\square$

\subsection{Off-chain $Z$-learning}

While off-policy $Q$-learning is not applicable to DSO, our earlier LMDP work carries over to the DSO framework more-or-less directly, and yields an off-chain DSO learning method. As in the L-MDP construction, the inputs to the optimization problem here are a baseline chain $\bar{p}(x'|x)$ and a state cost $r(x)$. Then for any chain $P$, the DSO cost is defined as:
\begin{equation}\label{LMDPcost}
L_L(x,P) \equiv r(x) + D_\textrm{KL}\left(P(\cdot|x) \parallel \bar{p}(\cdot|x)\right)
\end{equation}
When $P$ depends on parameters, those parameters affect $L$ only through $P$ and not directly; which is why we can write $L(x,P)$. Now suppose the ``parameters'' correspond to $P$ itself, i.e. we are optimizing over all Markov chains. This unconstrained problem has a unique global optimum, characterized as follows:

\begin{lemma}[Optimal non-parametric chain] Given baseline chain $\bar{p}(x'|x)$ and state cost $r(x)$, for any chain  $P(x'|x)$ define the DSO cost $L(x,P)$ as in (\ref{LMDPcost}) and the DSO episodic performance $J(P)$ as in (\ref{EPIperf}). For any function, define the linear operator:
\begin{equation}
G[f](x) \equiv \int \bar{p}(y|x) f(y) dy
\end{equation}
Then regardless of the initial state density $P_0(x)$, the global minimum of $J(P)$ over all chains $P$ is:
\begin{equation}\label{LMDP_pstar}
P^*(x'|x) = \frac{\bar{p}(x'|x) Z^\gamma(x')}{G[Z^\gamma](x)}
\end{equation}
where the function $Z(x)$ satisfies:
\begin{equation}\label{LMDP_zbel}
Z(x) = \exp(-r(x)) \thinspace G[Z^\gamma](x)
\end{equation}
\end{lemma}
\textbf{Proof.} Detailed proofs of this and related results (including uniqueness) are given in \cite{todorov2006} and subsequent papers in the LMDP series. Here we focus on the main derivation. At each $x$, think of $P(\cdot|x)$ as the action; resulting in an unusual MDP. The optimal value function for this MDP satisfies the Bellman equation:
\begin{equation}
\begin{aligned}
V(x) & = \min_{P(\cdot|x)} \left\{ L_L(x,P(\cdot|x)) + \gamma \int P(x'|x) V(x') dx' \right\} \\
& = \min_{P(\cdot|x)} \left\{ r(x) + \int P(x'|x) \left( \ln \frac{P(x'|x)}{\bar{p}(x'|x)} + \gamma V(x') \right) dx' \right\} \\
& = r(x) + \min_{P(\cdot|x)} \left\{ \int P(x'|x) \ln \frac{P(x'|x)}{\bar{p}(x'|x) \exp(-\gamma V(x'))} dx' \right\}
\end{aligned}
\end{equation}
The quantity being minimized is a KL divergence, except the density in the denominator is not normalized. The normalization factor is given by our linear operator: $G[\exp(-\gamma V)](x)$. Now we multiply both the numerator and the denominator by this normalization factor, and pull the redundant instance outside the integral (since it does not depend on $x'$). Define the function:
\begin{equation}
Z(x) \equiv \exp(- V(x))
\end{equation}
In this notation the normalization factor is $G[Z^\gamma](x)$ and the value function is $V(x) = - \ln Z(x)$. Then the Bellman equation for the optimal value function becomes:
\begin{equation}
-\ln Z(x) = r(x) - \ln G[Z^\gamma](x) + \min_{P(\cdot|x)} \left\{ D_\textrm{KL}\left( P(\cdot|x) \parallel \frac{\bar{p}(x'|x) Z^\gamma(x')}{G[Z^\gamma](x)} \right) \right\}
\end{equation}
The global minimum of KL divergence is zero, and is achieved when the two densities are equal -- which yields (\ref{LMDP_pstar}) in the lemma. Changing signs and exponentiating both sides of the Bellman equation (without the KL term which is now zero) yields (\ref{LMDP_zbel}) in the lemma. $\square$

In the first-exit undiscounted setting $\gamma = 1$, equation (\ref{LMDP_zbel}) is linear in $Z$, thus the name Linearly-solvable. A similar result is obtained in the average cost setting. There $Z$ is found as the principal eigen-function of $\exp(-r(\cdot)) \thinspace G[Z(\cdot)]$, with principal eigenvalue $\exp(-J)$ where $J$ is the optimal average cost per step.

The form of the globally optimal chain $P^*$ suggest how to design a parametric chain. Define a parametric function $Z(x,\theta) > 0$. Positivity can be guaranteed by setting $Z(x,\theta) = \exp(-E(x,\theta))$ for some energy-like function taking the place of the value function. Then define the (back to being parametric) DSO chain as:
\begin{equation}\label{LMDP_pfinal}
P_L(x'|x,\theta) \equiv \frac{\bar{p}(x'|x) Z(x',\theta)^\gamma}{G[Z(\cdot,\theta)^\gamma](x)}
\end{equation}
The corresponding DSO cost $L_L(x,\theta)$ is as in (\ref{LMDPcost}). The result (\ref{LMDP_zbel}) from the lemma can then be written as an expectation under the baseline chain:
\begin{equation}\label{LMDP_expect}
Z(x,\theta) = \exp(-r(x)) \thickspace \mathbb{E}_{x' \sim \bar{p}(\cdot|x)} \left[ Z(x',\theta)^\gamma \right]
\end{equation}
When $Z$ is allowed to be any function, this equation has a unique solution which yields the optimal chain. But here $Z$ is constrained by the parameterization, so there is no value of $\theta$ for which the equation is satisfied for all $x$; unless we are lucky and the unconstrained global optimum falls within our parametric family. This situation is similar to $Q$-learning, where convergence to the global optimum is guaranteed in the non-parametric setting, but once parameters are introduced it becomes an approximation. Applying the same form of approximation here yields our $Z$-learning algorithm.

The expectation in (\ref{LMDP_expect}) is under the baseline chain $\bar{p}(x'|x)$ and not the DSO chain $P(x'|x,\theta)$. So this is an off-chain method, but requires a specific chain. Using importance sampling, we can obtain a consistent algorithm sampling under any chain. Of particular interest however is the DSO chain itself:
\begin{equation}\label{LMDP_greedy}
Z(x,\theta) = \exp(-r(x)) \thickspace \mathbb{E}_{x' \sim P(\cdot|x,\theta)} \left[ \frac{\bar{p}(x'|x)}{P(x'|x,\theta)}Z(x',\theta)^\gamma \right]
\end{equation}
This is the analog of using the greedy policy in $Q$-learning. Indeed in our earlier LMDP work we observed that sampling from the latest $P(x'|x,\theta)$ converges much faster than sampling from $\bar{p}(x'|x)$.

The corresponding stochastic approximations can now be summarized as follows:
\begin{equation}
\begin{aligned}
\textrm{baseline:} \quad Z(x_t,\theta) & \leftarrow \exp(-r(x_t)) Z(x_{t+1},\theta)^\gamma, \quad x_{t+1} \sim \bar{p}(\cdot|x_t) \\
\textrm{greedy:} \quad Z(x_t,\theta) & \leftarrow \exp(-r(x_t)) \frac{\bar{p}(x_{t+1}|x_t)}{P(x_{t+1}|x_t,\theta)} Z(x_{t+1},\theta)^\gamma, \quad x_{t+1} \sim P(\cdot|x_t,\theta)
\end{aligned}    
\end{equation}
The arrows indicate targets for supervised learning of $\theta$. We omit the algorithmic details.

Greedy importance sampling here comes with extra computational cost. Substituting $P_L$ from (\ref{LMDP_pfinal}), the term inside the expectation in (\ref{LMDP_greedy}) equals the normalization factor:
\begin{equation}
\frac{\bar{p}(x'|x)}{P(x'|x,\theta)} Z(x',\theta)^\gamma = G[Z(\cdot,\theta)^\gamma](x)
\end{equation}
Therefore greedy $Z$-learning becomes:
\begin{equation}
Z(x_t,\theta) \leftarrow \exp(-r(x_t)) \int \bar{p}(y|x_t) Z(y,\theta)^\gamma dy, \quad x_{t+1} \sim P(\cdot|x_t,\theta)
\end{equation}
Note that the learning target at $x_t$ does not actually depend on the next state $x_{t+1}$. The update corresponds to in-place value iteration at each state we visit; indeed the last expression is identical to (\ref{LMDP_expect}). The integral can be approximated with a cubature formula. Alternatively, since this integral is itself an expectation, we can apply a double-sampling procedure: at each $t$ draw an intermediate sample $x_t' \sim \bar{p}(\cdot|x_t)$, set the learning target at $x_t$ to $\exp(-r(x_t)) Z(x_t',\theta)^\gamma$, and proceed as $x_{t+1} \sim P(\cdot|x_t,\theta)$.

The DSO parameterization we constructed here is of course also amenable to gradient descent, using Theorem 1 and Algorithm 1 to estimate the DSO gradient. In \cite{todorov2010} we considered this type of parameterization but with linear features for the underlying energy function: $Z(x,\theta) = \exp(-\theta^T \phi(x))$. After fitting a compatible value function approximation $\hat{V}(x,\omega) = \omega^T \hat{\phi}(x)$ using suitably modified features, we showed that the natural gradient reduces to $F(\theta)^{-1} \nabla_\theta J(\theta) = \theta - \omega$. This is yet another useful property of the LMDP framework that carries over to DSO.

\section{Broader applications of DSO}

In this final section we switch gears and address DSO applications beyond policy optimization, or rather, in conjunction with policy optimization. In the spirit of the times, we throw everything we might care about into a single objective and hit it with gradient descent. Our focus is on problems involving physical systems such as those arising in robotics and related areas. Applications to tuning of LLMs or other generative AI models are similar, except the states and actions become discrete.

Broadening the scope of DSO is achieved by packing multiple quantities in the parameter vector, which have different semantics and which correspond to solving different optimization problems. The DSO framework always treats $\theta$ as one vector, and the same general methods developed earlier are always applicable, regardless of what we packed into $\theta$. Here we consider the following parameter sets:
\begin{equation}
\theta = \left(\theta_\mu, \theta_f, \theta_\epsilon, \theta_h, \theta_\xi, \theta_c \right)
\end{equation}
These quantities in turn parameterize different functions summarized in the following table:

\begin{center}
\begin{tabular}{ l l }
 \hline
 prior: & $p_\theta(\theta)$ \\
 policy: & $a = \mu\left(x, \theta_\mu\right)$ \\
 dynamics: & $x' = f(x, a + \epsilon, \theta_f)$ \\ 
 dynamics noise: & $\epsilon \sim p_\epsilon(\cdot, \theta_\epsilon)$ \\
 measurement: & $y = h\left(x, \theta_h\right) + \xi$ \\
 measurement noise: & $\xi \sim p_\xi(\cdot, \theta_\xi)$ \\
 corrector: & $x = c\left(\tilde{x}, y, \theta_c\right)$ \\
 \hline
\end{tabular}
\end{center}

\noindent When designing DSO costs later, we will adopt a probabilistic approach where estimation-related errors are penalized by negative log-likelihood costs, and the parameters of the probability model itself can be learned. This is why the table contains three density models. The prior $p_\theta(\theta)$ is some overall prior which regularizes the optimization. It can be used for weight regularization in neural networks, as well as to keep physical parameters in reasonable ranges (so at to prevent negative mass estimates for example.)

The dynamics $x' = f(x, a + \epsilon, \theta_f)$ are based on some deterministic physics model whose action input is perturbed by the random variable $\epsilon \sim p_\epsilon(\cdot, \theta_\epsilon)$. The policy $a = \mu(x,\theta_\mu)$ is considered deterministic, with noise added to its output at the dynamics stage. The only source of stochasticity in the corresponding DSO chain is therefore the action noise. The reason for pushing noise through the dynamics, as opposed to adding it to the next state, is because physical systems usually have (soft or hard) constraints that are enforced by the dynamics. The resulting DSO chain is then:
\begin{equation}
P(x'|x,\theta) : \left\{x' = f\left(x, \mu(x, \theta_\mu) + \epsilon,\theta_f\right), \thickspace \epsilon \sim p_\epsilon(\cdot, \theta_\epsilon) \right\}
\end{equation}
In order to use the LMDP machinery, we also need a baseline chain corresponding to passive dynamics:
\begin{equation}
\bar{p}(x'|x,\bar{\theta}) : \left\{x' = f\left(x, \epsilon, \bar{\theta}_f\right), \thickspace \epsilon \sim p_\epsilon(\cdot, \bar{\theta}_\epsilon) \right\}
\end{equation}
Even though there is no policy here, $\bar{p}$ depends on the system parameters $\theta_f,\theta_\epsilon$. However, LMDPs do not allow $\bar{p}$ to depend on the current parameters being optimized. Thus $\bar{\theta}$ refers to some constants which are already fixed, perhaps from the previous DSO iteration.

Continuing with the table, we have the measurement model $y = h\left(x, \theta_h\right) + \xi$ which specifies how the measurements $y$ depend on the state $x$, and $\xi \sim p_\xi(\cdot, \theta_\xi)$ refers to sensor noise. This model has its own parameters which we may wish to calibrate -- sensor noise magnitudes, marker or camera placements, etc. Finally we have the corrector function $x = c\left(\tilde{x}, y, \theta_c\right)$ meant to be used in predictor-corrector estimation. Here $\tilde{x}$ is the predicted state, which is then corrected by the function $c$ given measurement $y$. Learning the parameters $\theta_c$ of this function (likely a neural network) corresponds to learning a state estimator.

Next we turn to cost function design. This will be done in modular fashion. First we describe the building blocks corresponding to common optimization tasks. Afterwards we explain how to combine them in order to solve specific (composite) problems.

\subsection{Cost function building blocks}

We design costs that depend on variables arising from the operation of the physical system -- states $x$, actions $a$, measurements $y$. In addition, we consider a dataset $\mathcal{D} = \{\bar{x}_t^n, \bar{a}_t^n, \bar{y}_t^n\}$ with recordings of (some of) these variables; in the same multiple-rollout format we adopted in Algorithm 1. The recorded data is treated as constant. We then design additional costs that depend on these constants.

\begin{equation}
\begin{aligned}
L_\textrm{prior}(\theta) & = - \ln p_\theta(\theta) \\
L_\textrm{clone}(\theta) & = - \tfrac{1}{N}\textsum_{n,t} \gamma^t \ln p_\epsilon\left(\bar{a}_t^n - \mu\left(\bar{x}_t^n, \theta_\mu\right), \theta_\epsilon\right) \\
L_\textrm{calib}(\theta) & = - \tfrac{1}{N}\textsum_{n,t} \gamma^t \ln p_\xi(\bar{y}_t^n - h(\bar{x}_t^n,\theta_h), \theta_\xi) \\
L_\textrm{sysid}(\theta) & = \tfrac{1}{N}\textsum_{n,t} \gamma^t \|\bar{x}_{t+1}^n - f(\bar{x}_t^n,\bar{a}_t^n,\theta_f)\|^2 \\
L_\textrm{hybrid}(\theta) & = \tfrac{1}{N}\textsum_{n,t} \gamma^t \|\bar{x}_{t+1}^n - f(\bar{x}_t^n,\mu(\bar{x}_t^n,\theta_\mu),\theta_f)\|^2 \\
L_\textrm{track}^{n,t}(x, \theta) & = \gamma^t \|\bar{x}_t^n - x\|^2 \\
L_\textrm{estim}^{n,t}(x, \theta) & = - \gamma^t \ln p_\xi(\bar{y}_t^n - h(x,\theta_h), \theta_\xi) \\
L_\textrm{Dmdp}(x, \theta) & = r(x, \mu(x,\theta_\mu)) \\
L_\textrm{Lmdp}(x, \theta) & = r(x) + D_\textrm{KL}\left( P(\cdot|x,(\theta_\mu,\theta_f,\theta_\epsilon)) \parallel \bar{p}(\cdot|x) \right)
\end{aligned}
\end{equation}

\noindent The first several costs in the table are purely data-driven and do not depend on the state. Optimizing any combination of such costs corresponds to supervised learning -- which is an easier problem and should be handled outside DSO. Technically it can also be handled within DSO, but then we will waste time computing the same gradient we could have obtained by differentiating $L(\theta)$ directly. The role of these data-driven costs here is to combine them with regular costs which depend on the state. We now provide explanations of what each cost building block is used for.

\begin{center}
\begin{tabular}{ l l }
cost & explanation \\
\hline
$L_\textrm{prior}(\theta)$ & Log-prior over parameters. Used for general regularization. \\
$L_\textrm{clone}(\theta)$ & Behavioral cloning cost. Learns a policy minimizing the difference between \\
& the actions and the policy outputs at the corresponding states.\\
$L_\textrm{calib}(\theta)$ & Sensor calibration cost. Learns a measurement model that explains the \\
& sensor measurements as a function of the corresponding states.\\
$L_\textrm{sysid}(\theta)$ & System identification cost. Used when (during data collection) the system is \\
& driven with known actions. Learns physics parameters which explain \\
& the observed state transitions. \\
$L_\textrm{hybrid}(\theta)$ & Hybrid of system identification and behavioral cloning. Used when only states are \\
& recorded but actions are not. Learns both the policy and the physics model. \\
$L_\textrm{track}^{n,t}(x, \theta)$ & Time-varying tracking cost. Used to specify time-varying goals / trajectories. \\
$L_\textrm{estim}^{n,t}(x, \theta)$ & Time-varying estimation cost. Used to learn a state estimator. \\
$L_\textrm{Dmdp}(x, \theta)$ & D-MDP cost for policy optimization. \\
$L_\textrm{Lmdp}(x, \theta)$ & L-MDP cost for policy optimization. \\
\hline
\end{tabular}
\end{center}

\subsection{Modeling with composite cost functions}

Next we describe several ways to combine the above building blocks, so as to formulate different optimization problems of interest. This list is not exhaustive, and neither is the table of building blocks. But it is sufficient to illustrate how broadly applicable DSO is.

\subsubsection*{Mechanism design}

The goal here is to design (or more likely, fine-tune) a physics model so that it can perform desired tasks; and simultaneously learn policies that make the model perform those tasks. An application is designing robots before building them. How do we know if our robot will be electro-mechanically capable of doing what we want? This is achieved in DSO by optimizing the policy and the physics parameters jointly, using a prior centered around an initial design and discouraging non-physical solutions.

\subsubsection*{Data-augmented policy optimization}

Policy optimization can spend a long time searching for a remotely-plausible policy (through random exploration) and then make more rapid progress once such a policy is found. One way to bypass this slow initial phase is to provide some demonstration data, and use it to initialize the policy via behavioral cloning. However we observed in \cite{rajeswaran2018} that performing behavioral cloning and policy optimization in parallel produces better results. In DSO this is achieved by simply adding the cloning cost and the policy optimization cost. Alternatively, if we only have example trajectories (say from videos), we can use the tracking cost instead. 

A related application is to augment policy optimization with the kind of data used for system identification and/or sensor calibration (via the corresponding data-driven costs.) This can help with sim-to-real transfer; and can be used instead of, or in conjunction with, domain randomization. Note that domain randomization itself is easily incorporated in DSO by using different chains $P^n$ in different rollouts.

\subsubsection*{Learning a state estimator}

This application is somewhat different and we need to modify the DSO chain as follows:
\begin{equation}
P_t^n(x'|x,\theta) : \left\{x' = c\left(f\left(x, \bar{a}_t^n + \epsilon,\theta_f\right), \bar{y}_t^n,\theta_c\right), \thickspace \epsilon \sim p_\epsilon(\cdot, \theta_\epsilon) \right\}
\end{equation}
These dynamics combine a prediction step under the physics model $f$, followed by a correction implemented by the corrector $c$. Now we apply the estimation cost which penalizes measurement prediction errors. Learning in this setting corresponds to jointly optimizing the corrector parameters, the measurement model parameters, and the physics model parameters. Of course some of them can be fixed and not included in the parameter vector $\theta$ handed to DSO; while their values are still used to evaluate the system models.

Note that here we are learning a fixed corrector which can then estimate states at runtime. Traditional state estimation, as well as the dual problem of trajectory optimization, involve decision variables that are functions of time  (states and actions.) These problems do not fit in DSO. Nevertheless, the corresponding optimizers can be used in conjunction with DSO -- to generate synthetic data which goes into the dataset $\mathcal{D}$ and thereby facilitates DSO through data-driven costs.

\subsubsection*{Learning brain-body dynamics}

We finish with an unusual yet promising application. Suppose our policy, or state estimator, or system identifier, is implemented as a recurrent neural network. This is broadly related to \cite{ha2018} except there the focus was on learning the system/world model with a recurrent network and putting a simple static policy on top of it; while here we leverage the power of recurrent networks for control-related computations, after a system model is available. The unit activations in the network correspond to a new dynamic state $x_g$ which we combine with the physics state $x_f$; thus augmenting the DSO state vector as $x = (x_f, x_g)$. The physics and the network dynamics are updated synchronously at each time step. To connect the network to the rest of the system, we need input weights mapping from physics states at the previous time step; recurrent weights which point forward in time (we are unfolding the network dynamics, similar to backpropagation-through-time;) and output weights mapping from network states to actions that drive the physics. We also allow direct weights that bypass the network and implement a linear feedback controller. All these new weights are included in $\theta$. The recurrent dynamics update can be fully determined by its input (which of course depends on its previous state) or it can be partly autonomous as in LSTM.

Putting all this together, we now have brain-body dynamics:
\begin{equation}
\begin{aligned}
x_f' & = f(x_f, W_D(\theta) x_f + W_O(\theta) x_g + \epsilon_f) \\
x_g' & = g(x_g, W_I(\theta) x_f + W_R(\theta) x_g + \epsilon_g)
\end{aligned}
\end{equation}
$W_I, W_R, W_O, W_D$ extract the corresponding components of $\theta$ and reshape them into weight matrices: Input, Recurrent, Output, Direct. The network dynamics $g$ are normally applied component-wise for each unit, but could also implement pooling. We can further express this jointly as: $x' = \tilde{f}(x, W(\theta)^T x + \epsilon)$. The latter form highlights the fact that a linear policy parameterization is likely sufficient here -- because the network should be able to generate a rich set of dynamic features, reminiscent of kernel methods. This entire system is naturally expressed as a DSO, and handled with our methods.

\section{Conclusion}

We developed a general framework for dynamical system optimization. It solves control-related problems without reference to controls or actions, while inheriting and indeed simplifying many of the tools from approximate Dynamic Programming and Reinforcement Learning. It also merges our earlier LMDP framework with traditional MDP formulations, and unifies seemingly disparate concepts. The present paper is entirely theoretical. We look forward to implementing and testing these algorithms in future work.

\subsection*{Acknowledgments}

Thanks for Yuval Tassa, Aravind Rajeswaran and Hilario Tome for comments on the manuscript.

\end{document}